\documentclass[11pt]{article}
\usepackage[margin=1in]{geometry}
\usepackage{booktabs}
\usepackage{array}
\usepackage{microtype}
\usepackage{parskip}
\usepackage{titlesec}
\usepackage{xcolor}
\usepackage[hidelinks]{hyperref}
\usepackage{helvet}
\usepackage{tikz}
\usetikzlibrary{positioning,arrows.meta,decorations.pathreplacing}
\titleformat*{\section}{\large\bfseries}
\titleformat*{\subsection}{\normalsize\bfseries}

\title{\textbf{Auditing Long-Term Memory Evaluation:\\Repeated Judging, Reader Variation,\\and Negative Controls}}
\author{Christopher J. Chanhnourack\thanks{Correspondence: \texttt{cj@centennialsystems.com}, Centennial Defense Systems. Evidence release: \url{https://github.com/cjchanh/longmemeval-evidence}. Source-tree evidence commits: 477726726 \textperiodcentered{} 04eb64763 \textperiodcentered{} ad71321ea \textperiodcentered{} ba805e575 \textperiodcentered{} 08e5784bb \textperiodcentered{} ad43cc99a \textperiodcentered{} 9b0711c39.}\\Centennial Defense Systems (Archivist)}
\date{Revised technical report --- October 1, 2026\\{\small Version 2; replaces arXiv:2609.38021v1}}

\begin{document}
\maketitle

\begin{abstract}
\noindent This report audits evaluation of a long-term-memory retrieval chain on the 500 LongMemEval-S development questions. Its strongest historical reader lane scores 479 and 475 under an adapted GPT-4o rubric; re-judging the same pass-1 answers changes three labels and yields 478. Fixed-answer knowledge-update re-scoring gives 70/72 under the upstream template and 69/72 under the modified template. Reader lanes span 93 to 479 on fixed packets; paired tests between the two strongest historical lanes establish neither superiority nor equivalence. A different-family reader, configured without client tools or operator files, scores 474, 1.0 percentage point below the headline pass (paired 95\% interval [-3.0,+1.0]). Live reader request bodies were not retained. With the same requested reader label, route and judge snapshot, the full package scores 474 versus 454 for baseline sessions, a difference of +4.0 percentage points [95\% interval +2.2,+6.0]. Eighteen of the 23 gains, and no losses, occur where baseline packets lacked listed evidence; this post-hoc split does not identify a component effect. In recovered LoCoMo data, token-F1 gains do not survive answer-line extraction. A negative control rejects a verifier that repairs three wrong drafts but breaks eleven correct ones. All questions were used to develop the components; no untouched holdout was evaluated. These findings do not establish a new leaderboard leader or transferable memory advantage. The A/D comparison has one pass per arm, including six reused identical-prompt outcomes, with no pinned reader snapshot; B/C and repeats remain unrun. Original headline requests cannot be reconstructed and stages 1--4 remain closed. Released artifacts support packet inspection and saved-verdict recounting and re-scoring; they do not reconstruct the method.
\end{abstract}

\section{Introduction}
A chat assistant with long-term memory has to answer questions about the
user's own history --- \emph{how many different doctors did I visit?} ---
when the evidence is scattered across dozens of past conversations recorded
on different days, sometimes restated and sometimes superseded.
LongMemEval-S (Wu et al., 2024) measures exactly this: 500 questions, each
over a history of roughly 40--60 sessions. The failure that matters is
quiet: a system that never retrieves the one session where a fact was
stated answers fluently and wrongly (\S3.6 traces a real case). In settings
where an answer has to be checked after the fact, the question is not only
whether the system was right but whether anyone can see why.

We built a retrieval chain in which every stage before the final answer is
deterministic code --- hybrid candidate retrieval, cross-encoder reranking, a packet
compiler, and mechanical reasoning scaffolds --- and a language model appears
only once, as a replaceable reader (Figure~1). Because the stages below the
reader are fixed, the same packets can be handed to any reader, and every artifact from the packet onward can be released and checked.

Long-context memory benchmarks are typically reported as single numbers from
single runs, judged by an LLM, with no disclosure of run-to-run or judge
variance. At the top of a leaderboard, where entries are separated by a handful
of questions out of 500, this practice makes the ranking partially an artifact
of measurement noise.

This report contributes an inspectable evaluation case study. Its main result is an observed range of 475--479 correct answers across two passes, accompanied by repeated judging, per-question outcomes, and negative controls. The range contains the published Chronos count of 478; it is the range of two observed runs, not a confidence interval or an estimate of future performance. Differences in model generation, prompt, data version, and runtime context prevent a controlled comparison with Chronos.

Three observations motivate the release. First, re-judging unchanged answers flips three labels, so a one-answer ranking margin is smaller than the disagreement observed in this run. Second, scores vary widely across reader lanes on fixed packets, making the reader configuration part of the object being evaluated. Those lanes also differ in tools and additional context, so their spread does not isolate a pure model effect. Third, a verifier that helps on selected failures can damage more initially correct answers than it repairs. Publishing the rejected intervention makes that failure checkable.

The retrieval components follow established retrieve-and-rerank practice. The contribution is the documented evaluation procedure and its failure cases, rather than a claim that this combination is a new retrieval primitive. Development used all 500 LongMemEval-S questions. A later LoCoMo investigation, recovered for this revision, found no overall answer-line token-F1 gain distinguishable from zero (\S5.5). It does not provide a pristine holdout or a scaffold-specific ablation. The paired knowledge-update comparison in \S5.6 now measures one rubric difference on fixed answers. A pinned reader with complete request capture and an exposure-audited transfer set remain necessary.

\textbf{Terms used below.}
\begin{itemize}\setlength{\itemsep}{1pt}\setlength{\parskip}{0pt}\setlength{\parsep}{0pt}
\item \emph{Pass, pair.} A pass is one full run of all 500 questions through a reader and the adapted GPT-4o judge (\S2); a pair is two passes of the same configuration (\S4.1).
\item \emph{Flip row.} A question whose verdict differs between the two passes of a pair, or between two judgings of the same answer.
\item \emph{Gold, gold-complete.} Gold is the answer key: the expected answer and the sessions that hold its evidence. A packet is gold-complete when it contains every gold session.
\item \emph{Proxy token.} A proxy-token count is the character count divided by four, rounded up; packet budgets use it.
\item \emph{Scaffold, operators.} A scaffold is the deterministic evidence index compiled from a packet (\S3.4); operators are its versioned extraction rules (v3.3, v3.4).
\item \emph{CLI lane.} The headline Opus reader's route: the Claude Code command line, with the built-in tools disabled (\S3.5).
\item \emph{Dossier, ex-dossier.} The dossier is the row-by-row list of eight questionable answer-key rows chosen from the grok pair's failures, one strict defect and seven boundary cases (\S6.3); ex-dossier scores leave those rows out (492 questions).
\item \emph{Closed-book.} A reader that sees only its prompt, with every tool turned off.
\end{itemize}
The system under test is the retrieval core of Archivist, a local-first conversational memory product. The release supports inspection and re-scoring of materialized packets, scaffolds, reader outputs, judge verdicts, and control receipts. Stage sources remain held (\S8), and the LoCoMo diagnostic does not establish a transferable memory advantage (\S5.5, \S7).

\subsection{Related work}
Agentic memory systems (MemGPT, Packer et al., 2023; production systems such as
Mem0, Chhikara et al., 2025, and Zep, Rasmussen et al., 2025) focus on memory \emph{stores} and management policies; our
contribution is orthogonal --- determinism below the reader, with the store
fixed. Long-term memory benchmarks include LongMemEval (Wu et al., 2024), used
here, and LoCoMo (Maharana et al., 2024); the main study concerns LongMemEval-S, with a recovered LoCoMo diagnostic in \S5.5 and no claim of cross-benchmark generality (\S7). The retrieval stack follows the
retrieve-and-rerank tradition (Nogueira \& Cho, 2019; BGE-M3 family, Chen et
al., 2024); reciprocal rank fusion (Cormack et al., 2009) is among the
extension-ordering baselines swept in \S3.3. On measurement: a
self-consistency step (Wang et al., 2023) was built and later withdrawn when
internal review showed its two-pass framing was a derived identity (\S4.4); the judge-variance study operationalizes known LLM-as-judge concerns
(Zheng et al., 2023); the two-pass reporting rule answers the single-run
critique of Dodge et al.\ (2019).

\section{Benchmark and evaluation protocol}
\textbf{LongMemEval-S} (Wu et al., 2024) contains 500 questions over long
multi-session chat histories: 470 answerable questions across six types
(single-session-user, single-session-assistant, single-session-preference,
multi-session, temporal-reasoning, knowledge-update) and 30 abstention
questions (\texttt{\_abs}) whose correct behavior is to decline to answer.
Each question carries a haystack of $\sim$40--60 sessions with timestamps;
evidence for a question may be spread across several sessions recorded
weeks apart. We ran the cleaned release of the file (\texttt{longmemeval\_s\_cleaned}, the September 2025 cleanup of the history sessions that the benchmark's maintainers describe as preventing interference on answer correctness). Chronos does not say whether it used the cleaned or the original file, so comparisons with its 478 may also cross a data-version difference.

\textbf{Judge.} We use an adapted version of the benchmark's protocol, GPT-4o
(\texttt{gpt-4o-2024-08-06}) with the benchmark's rubric templates, with one deviation. Our templates are SHA-256 pinned
(\texttt{9c9d67fab129\ldots}) and released verbatim beside a pinned copy of the benchmark's \texttt{evaluate\_qa.py} and a test that compares them: the preference and abstention templates match it exactly, and the base and temporal-reasoning prompts differ only by a trailing space. The knowledge-update prompt differs in substance: ours appends the update rule to the base template, so it keeps two sentences that the official knowledge-update template omits (a response equivalent to the answer counts as correct; one with only a subset of the required information does not). The 72 answerable knowledge-update rows of every historical run were judged with that prompt. Section 5.6 compares both templates on the saved Opus answers in three new repetitions; other reader lanes retain the original modified-template judgments. Throughout this report, the historical GPT-4o scores use this adapted rubric. An exact-upstream result must be labeled separately; model identity alone does not make these judgments official-template measurements. Every scoring run executes live positive and
negative judge-control verdicts against a 0.9 pass threshold (preference rows
are excluded from the positive control arm because their gold is a rubric, which a correct judge does not accept as a response); every historical control arm described in the original study passed at 100\%; the new controls are reported in \S5.6. Adapted-judge full runs used 12 positive and 12 negative controls (19 and 20 for the pair run on the earlier scaffold file \texttt{facts.jsonl}, \S5.3); each run's receipt records its own counts. The
harness records the resolved model snapshot per verdict and fails closed on
authentication errors.

\textbf{Selected historical comparison points (author-reported; not a current or exhaustive leaderboard).} Chronos and Mastra sources are pinned in the release; the other observations retain the dates and limitations described in \S8. The pinned Chronos High count is \textbf{478/500 (95.60\%)} (PwC,
arXiv 2603.16862, Mar 2026; PDF sha256 \texttt{a6a75d61\ldots}; generator Claude Opus 4.6, LongMemEval judge
protocol, per-category table published --- their Table~2 and Appendix~A,
Table~4). Two widely-quoted higher percentages
are not raw-comparable: Mastra OM's ``94.87\%'' is a category-unweighted
average --- their own per-category counts sum to \textbf{468/500 raw}
(generator gpt-5-mini; page sha256 \texttt{cd0bf092\ldots}) --- and OMEGA's ``95.4\%'' is likewise task-averaged
with raw 466/500, graded by GPT-4.1 rather than the canonical judge (\url{https://omegamax.co/benchmarks}). 

A higher
claim (481/500, ``agentmemory V4'') appears in a public
repository (\url{https://github.com/JordanMcCann/agentmemory}). Its published runner, run log and result file name \texttt{longmemeval\_oracle.json} as the input, the benchmark's variant that holds only the evidence sessions (1.9 per question on average, against roughly 40--60 in LongMemEval-S), so we treat its retrieval condition as different from S. This is an input-condition distinction, not a verdict on the method; it also means the largest reported number does not establish a like-for-like leaderboard ordering.

Agent Zero
Memory (arXiv 2608.29606, Aug 2026) reports the same 95.60\% on 500
LongMemEval questions without naming the variant or the judge, and its
comparison table omits Chronos; we cannot classify it as a like-for-like S result from that description. Reader tier
varies freely across published entries (GPT-4o through Opus 4.6; our headline
reader's alias resolved to Opus 5 on a same-day probe, \S5.3); we follow the
field convention of reporting our strongest configuration alongside the full
reader ladder.

\section{System architecture}
The chain has five stages. Stages 1--4 are deterministic: byte-identical
outputs across runs are checked by verification gates, not assumed (verified by the author on the pinned environment; the stages are not released, so this is not independently reproducible; hardware and library versions are recorded in the
release).

\begin{figure}[htbp]
\centering
\footnotesize
\begin{tikzpicture}[
  box/.style={draw, rounded corners=2pt, align=center, text width=2.75cm, minimum height=1.9cm, inner sep=3pt},
  llm/.style={box, fill=black!8},
  arr/.style={-{Stealth[length=2mm]}, thick},
  node distance=0.28cm]
\node[box] (dense) {\textbf{1 Retrieve}\\[3pt] {\scriptsize three lanes}\\ {\scriptsize all gold in pool:}\\ {\scriptsize 468/470}};
\node[box, right=of dense] (ce) {\textbf{2 Rerank}\\[3pt] {\scriptsize cross-encoder}\\ {\scriptsize session scores}\\ {\scriptsize \phantom{x}}};
\node[box, right=of ce] (pack) {\textbf{3 Compile}\\[3pt] {\scriptsize packet $\le$16 sessions}\\ {\scriptsize gold-complete:}\\ {\scriptsize 462/470}};
\node[box, right=of pack] (fact) {\textbf{4 Scaffold}\\[3pt] {\scriptsize evidence index}\\ {\scriptsize for 240 of 470}\\ {\scriptsize questions}};
\node[llm, right=of fact] (read) {\textbf{5 Read (LLM)}\\[3pt] {\scriptsize replaceable reader}\\ {\scriptsize eight readers}\\ {\scriptsize measured}};
\draw[arr] (dense) -- (ce); \draw[arr] (ce) -- (pack); \draw[arr] (pack) -- (fact); \draw[arr] (fact) -- (read);
\draw[decorate, decoration={brace, amplitude=5pt}] ([yshift=5pt]dense.north west) -- node[above=6pt] {deterministic code, verified by receipts and negative controls} ([yshift=5pt]fact.north east);
\end{tikzpicture}
\caption{The chain. Stages 1--4 are deterministic code; only stage 5 calls a
language model, and it can be swapped without changing the packets it
reads. The adapted GPT-4o judge scores the reader's answer. Counts are from
\S3.1--\S3.5.}
\end{figure}
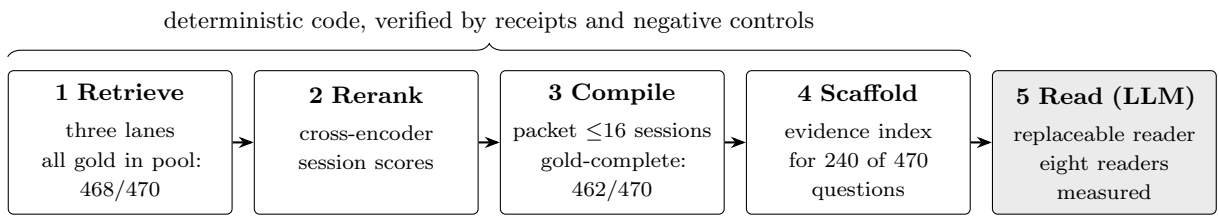

\textbf{3.1 Candidate retrieval.} Three lanes nominate sessions from the haystack,
and their union is the candidate pool: dense embedding retrieval, keyword
full-text search, and a bounded semantic-expansion lane that admits further
sessions by embedding similarity (it only adds candidates; it never removes or
reorders them). All inference is local. Pool ceiling: 468/470 questions have all
gold sessions somewhere in the pool; of the gold sessions in the released answerable packets (gold labels from the public dataset), 10 arrive through the semantic-expansion lane alone.

\textbf{3.2 Cross-encoder reranking.} A cross-encoder
(\texttt{BAAI/bge-reranker-v2-m3}) scores each candidate session against the
question; session score is the maximum chunk score (chunk = date header +
turn). Scoring is rank-deterministic (fp16, fixed batch, tie-break by prior
rank).

\textbf{3.3 Coverage-first packet compilation.} For each question a reading
packet is compiled under a hard budget (16 sessions / 40k proxy tokens): the
dense baseline top-10 is preserved as a byte-identical prefix,
remaining slots are filled in cross-encoder order with term-preserving
extractive compression, and budget overruns drop the lowest-ranked extension
first. The compiler is validated by determinism (two compiles
byte-identical), a gold-permutation negative control (shuffled gold answers
must not change output), a question-ID sabotage control (renamed IDs must
not change output), and budget accounting. Packets are gold-complete for 462/470 questions; a sweep of
all five extension-ordering policies found the shipped ordering best of the five (462 vs.\ 454 for
the next best), measured on these same questions.

\textbf{3.4 Deterministic reasoning scaffolds (``facts'').} For questions
whose type benefits from mechanical reasoning (240 of 470), the chain emits
a deterministic evidence index computed from the packet: session chronology,
a canonical user-statement recency domain with clock-granular tie keys, a
value-preference invariant (statements carrying candidate values outrank
bare recency), candidate enumeration
with quantity roll-up for counting questions, and granularity/unit and
event-anchor rules (contract \texttt{v3.4-facts-20260831}). Scaffolds are code
with in-code coupling invariants that raise on internal inconsistency. The
scaffold reads only the packet and the question, so it is designed to be
gold-blind and question-ID-independent; its source is held (\S8). Scaffold rule
sets are versioned as \emph{operators} (v3.3, v3.4); the ladder in \S5.3 moves
between them.

\textbf{3.5 Replaceable reader.} The packet, scaffold, and question go to an
LLM reader. The reader is deliberately replaceable; we measure the chain
under eight readers (\S5.4). The headline Opus reader is called through the Claude Code command line (the
``CLI lane'') with the built-in tools disabled; \S7 audits every lane, including the operator context this one received. The grok-4.6 reader
runs at two reasoning-effort settings, \emph{high} and the maximum, \emph{xhigh};
the xhigh runs used an agentic command-line transport. The headline pair uses the Claude Opus reader
over the CLI lane (\S5.1); \texttt{grok-4.6-high} and GLM-5.3 are the
secondary readers (\S5.4).

\textbf{3.6 A worked example.} One released question, traced through every
stage. The question and its haystack are in the public dataset; the packet,
scaffold, reader outputs and verdicts are in the release.

\begin{center}
\small
\renewcommand{\arraystretch}{1.15}
\begin{tabular}{>{\raggedright\arraybackslash}p{2.3cm}>{\raggedright\arraybackslash}p{12.4cm}}
\toprule
Stage & Question \texttt{gpt4\_f2262a51}\\
\midrule
Question & ``How many different doctors did I visit?'' (asked 2023/05/30).
Gold: three --- a primary care physician, an ENT specialist, a dermatologist.\\
Haystack & 45 sessions, 479 turns, about 78,000 words. The evidence sits in
three sessions from May 20--22; Dr.~Patel is named in two of them.\\
1 Retrieve & Neither the dense lane nor the keyword lane nominates any of the
three evidence sessions; the semantic-expansion lane does. The flat-reading baseline, which read ten dense-retrieved sessions,
answered that no doctor visits were recorded (judged wrong).\\
2--3 Rerank and compile & The dense top-10 is kept as the packet's first ten slots;
slots 11--16 are filled from the rest of the pool in cross-encoder order, and the
three evidence sessions land at slots 13, 15 and 16. Packet: 16 sessions, 28,544 of 40,000 proxy tokens.\\
4 Scaffold & The counting operator lists the packet's sessions in date order
and enumerates nine candidate phrases, each marked
as a candidate, not a confirmed item --- including irrelevant ones, such as a
passport renewal.\\
5 Reader & The Opus reader names Dr.~Smith (primary care), Dr.~Patel (ENT)
and Dr.~Lee (dermatology), treats the second Dr.~Patel mention as the same
doctor, sets aside a colonoscopy that was only being considered, and answers
3. Correct in both Opus passes and both grok passes.\\
\bottomrule
\end{tabular}
\end{center}

The baseline's error was a retrieval error: the evidence was never in its
context, so no reader could have answered from it. The recovery came from a wider
candidate pool, ordered by the cross-encoder. The example is also an unusual one:
across the release, only 10 gold sessions reach a packet through the
semantic-expansion lane alone, and three of them are this question's. The example also shows
what the scaffold does not do. It enumerates candidates mechanically and
leaves the judgment --- is a planned colonoscopy a visit? is Dr.~Patel one
doctor or two? --- to the reader, which is where the remaining errors live
(\S6.1).

\section{Measurement methodology}
\textbf{4.1 Two-pass rule (fixed before the Opus pair).} A nondeterministic reader means a
single full run at a one-point margin is not a reportable number. Our
promotion rule (its earliest record in the release is dated 2026-08-31T06:20Z; the Opus pair was scored from 2026-09-01T03:12Z): a headline
requires two full
passes agreeing, or a stated spread when the pair straddles the bar, with no clean-beat headline. The $\pm$ we report is half the range between the two passes; the flip-row counts of \S5.1 show the run-to-run noise.

\textbf{4.2 Judge disagreement is observed, not assumed.} We re-judged both passes
of the headline Opus pair with a second judge, GLM-5.3 (\texttt{glm-5.3:cloud}), on identical rows: agreement 493/500 (98.6\%) in each pass, and the second judge scores the pair 472/472. In pass 1 all seven disagreements are adapted-judge-only credits, five of them abstention rows; in pass 2 five are adapted-judge-only and two are second-judge-only. An earlier study re-judged both passes of the earlier grok-4.6-high pair (471/470, scaffold file \texttt{facts.jsonl}, \S5.3) with \texttt{glm-5.3-flash}: agreement
98.2\%/97.8\%, the second judge harsher on net. We also observed the adapted GPT-4o
judge returning opposite verdicts on a byte-identical answer across runs. At
these margins the observed judge disagreement limits ranking claims; we log such rows rather than
resolve them by re-rolling.

\textbf{4.3 Negative controls at the improvement layer.} Every proposed
improvement runs against a 60-row control set of baseline-correct questions (58 of them correct in both passes of the grok-4.6-high pair, the incumbent when the gate ran) before integration. One stage failed it: a verifier stage that recovered 3
previously-wrong rows was \emph{rejected} because the control replay showed it
flipped 11 of 58 correct drafts to wrong (\S6.2). The frozen acceptance rule
--- zero control flips --- blocked it from ever touching the headline, while
the xhigh reader variant (\S5.1) passed that same 60-row control with zero
flips before its full pair was launched.

\textbf{4.4 Self-consistency tie-break (built, then withdrawn).} The raw
pair's spread came from eight verdict-flip rows, and a tie-break rule frozen
before any additional read gave each flip row three more reads with a
five-vote majority. The resulting number was withdrawn from any claim:
internal review showed the construction wrote one derived verdict into both
passes (an identity, not two agreeing measurements) and its flip-row
selection was judge-conditioned. The artifacts remain released as a record;
no self-consistency number is claimed anywhere in this report.

\textbf{4.5 Development and reporting boundaries.} Inference used generic prompts without question-ID-specific branches. Generic rules were revised after inspecting particular question failures; this is development-set tuning. Re-judgings of reported results, the stale-verdict repair of one xhigh row, and the full reproduction re-judge are disclosed in \S5.1 and \S8. Superseded partial judgings remain in the release. Failed routes are disclosed in \S6, and the recovered LoCoMo study omitted from v1 is added in \S5.5. Counterfactual gold-defect adjudications do not change any headline score.

\section{Results}
\subsection{Repeated LongMemEval-S measurements}
\begin{center}
\begin{tabular}{lcc}
\toprule
Configuration & Pass 1 & Pass 2\\
\midrule
\textbf{v3.4, Claude Opus reader (CLI lane)} & \textbf{479/500} & \textbf{475/500}\\
v3.4, grok-4.6 high (raw pair) & 476/500 & 474/500\\
v3.4, grok-4.6 xhigh (agentic transport) & 461/500 & 465/500\\
\bottomrule
\end{tabular}
\end{center}

Dossier exclusions and counterfactual adjudications are reported only in Appendix~\ref{app:dossier}; they are not alternative benchmark scores.

The Opus pair is the strongest observed lane: the pair's both-pass-stable count is 473 (vs.\ the grok pair's 471);
pass 1 is perfect on three of six answerable types (single-session assistant,
user, preference). Its 8 flip rows (6 lost --- including 2 abstentions and 1
dossier-listed flip-prone row --- 2 gained) match the flip-noise structure of
the grok pair (also 8 rows; the earlier grok pair on the \texttt{facts.jsonl} scaffolds shows 15 and the regressed xhigh pair 20).

Under the two-pass rule (\S4.1) the pair reports as
\textbf{477$\pm$2} (half the range between the passes), bracketing the published 478; neither margin exceeds the three-label disagreement observed in one re-judge of pass 1. A pre-release re-judge of pass 1 under the same adapted GPT-4o judge
returned 478 --- three verdict flips on identical text (\S8) --- so the
one-point margin is smaller than the observed three-label disagreement. This single re-judge does not establish a universal noise floor or estimate the full judging distribution.

Judge controls 12/12 and 12/12 in both passes; a 25-row probe preceded the pair
(11/25 recovered, 8 of them stable-wrong for the grok pair).

The xhigh variant is a measured regression ($-15$/$-9$), reported in full. Its sessions also looked outside their prompts, some of them for the answer; \S7 reports what they reached and shows that the regression survives removing every row whose answering session did so.
This changes effort and agentic transport together, so the regression cannot be attributed to reasoning effort alone.
Its gate evidence had looked favorable --- a 25-row probe on the high pair's
wrong set recovered 9, and a 60-row negative control on baseline-correct rows
showed zero flips --- but the full pair exposed the gates' limits: only 4
probe recoveries held, and the true correct-row flip rate ($\sim$3.4\%) was
small enough that a 60-row control had a $\sim$13\% chance of showing zero
flips. Row-level attribution of pass 1's 20 losses: 4 are judge flips on
a final answer identical to the high run's; the other 16 are reader losses, 15 of which share one over-skepticism
signature --- wrong abstentions on answerable rows, over-filtered counts,
hedged golds. In the xhigh configuration these losses look like the reader second-guessing correct evidence; effort, transport and tool access changed together, so the cause is not identified. (One v1 verdict was a staleness artifact, not a reader loss: the
reader resumed one row eleven minutes after the judge had scored it on an
empty response; a surgical v2 re-judge under fresh 12/12 controls restored
it, $460 \to 461$. The authoritative repaired file is
\texttt{full-v34-xhigh\_pass1/judge\_gpt4o\_v2/rescored.jsonl}; the unsuffixed directory retains the stale 460. The sole changed label is reproduced in the revision supplement. The high run also missed that row, so the repair adds a gain and removes no loss.)

Raw grok-pair detail: answerable 450/470 and 447/470; abstention 26/30
and 27/30. Judge controls passed in every scoring run. The strict-defect counterfactual is confined to Appendix~\ref{app:dossier}.

\textbf{Paired comparisons within the released runs.} The release supports exact two-sided McNemar tests because its verdicts can be joined by question ID. Opus versus Grok has 13 Opus-only and 10 Grok-only correct answers in pass 1 ($p=0.677639$), and 12 versus 11 in pass 2 ($p=1.000000$). Opus pass 1 versus pass 2 has 6 versus 2 discordant answers ($p=0.289063$). These are post-hoc comparisons of adapted-judge binary outcomes on the development questions. They do not test equivalence, establish that the underlying models are interchangeable, remove selection bias, or isolate differences in runtime context. The missing external per-question verdicts prevent the corresponding test against Chronos.

\subsection{Per-question-type breakdown}
\begin{center}
\small
\begin{tabular}{lccccc}
\toprule
Question type & $n$ & Raw p1 & Raw p2 & xhigh p1 / p2 & Opus p1 / p2\\
\midrule
single-session-assistant & 56 & 56 (100.0\%) & 56 (100.0\%) & 55 / 55 & 56 / 56\\
single-session-user & 64 & 63 (98.4\%) & 63 (98.4\%) & 61 / 60 & 64 / 64\\
knowledge-update & 72 & 70 (97.2\%) & 71 (98.6\%) & 69 / 70 & 69 / 70\\
single-session-preference & 30 & 29 (96.7\%) & 29 (96.7\%) & 25 / 25 & 30 / 28\\
temporal-reasoning & 127 & 122 (96.1\%) & 122 (96.1\%) & 121 / 123 & 123 / 123\\
multi-session & 121 & 110 (90.9\%) & 106 (87.6\%) & 103 / 106 & 110 / 109\\
abstention & 30 & 26 (86.7\%) & 27 (90.0\%) & 27 / 26 & 27 / 25\\
\midrule
\textbf{Total} & \textbf{500} & \textbf{476} & \textbf{474} & 461 / 465 & \textbf{479 / 475}\\
\bottomrule
\end{tabular}
\end{center}
{\footnotesize Judge: the adapted GPT-4o judge; the 72 knowledge-update rows use the modified prompt (\S2). ``Raw'' columns are the grok-4.6-high pair.}\par\smallskip

Five of six answerable types are pass-stable to within one question; the raw
pair's entire answerable spread is one type (multi-session), partially offset
by knowledge-update. Against the pinned Chronos category table on its own abstention-folded
basis ($n$: KU 78, MS 133, SSA 56, SSP 30, SSU 70, TR 133 --- their
Appendix~A), our pass-1 deficit is $-3$ knowledge-update (75/78 vs.\ their
78/78; our knowledge-update judge prompt departs from the official one, \S2, and the upstream template recovers one of these three on the saved answers, \S5.6), offset by $+2$ multi-session (120/133 vs.\ 118/133), $+1$
single-session-user (70/70 vs.\ 69/70) and $+1$ temporal-reasoning (128/133
vs.\ 127/133), with assistant and preference level --- net $+1$, which is 479
vs.\ their 478. Pass 2 carries the same $-3$ knowledge-update and the same
$+1$ single-session-user and $+1$ temporal-reasoning, ties on multi-session
(118/133), and loses $-2$ on preference (28/30) --- net $-3$,
which is 475 vs.\ 478. Our multi-session count is therefore 2 higher in pass 1 and equal in pass 2, which at $n=133$ and with the comparability limits of \S2 is not evidence of a difference; the pass-to-pass difference is split between
multi-session and preference ($-2$ each).

\subsection{Build history}
\begin{center}
\begin{tabular}{lc}
\toprule
Stage & Score /500\\
\midrule
Dense retrieval + flat reading (Flash reader, v2 prompts) & 452/499$^{1}$\\
+ chain, scaffold set \texttt{f3value} (before v3.3), GPT-5.6 Sol reader & 468\\
+ cross-encoder rerank (v3, grok reader) & 473\\
+ \texttt{facts.jsonl} scaffolds (label v3.3), grok high & 471 / 470\\
+ v3.4 operators (adapted-judge pair, grok high) & 476 / 474\\
+ reader xhigh + agentic transport & 461 / 465 (regression)\\
\textbf{+ reader tier: Claude Opus, same substrate} & \textbf{479 / 475}\\
\bottomrule
\end{tabular}
\end{center}
{\footnotesize Single ruler: the adapted GPT-4o judge, knowledge-update prompt excepted (\S2).}\par\smallskip
Rungs change more than one component at a time (reader, operators, prompts), so
the ladder is a build history, not a component-wise ablation. Scaffold versions follow the per-row scaffold hashes in the release, not the run labels: the 471/470 pair carries the prompt label v3.3 but read the earlier \texttt{facts.jsonl} (generated 2026-08-30T19:05Z; v3.3 was generated 2026-08-31T00:25Z), and the 468 Sol run read the \texttt{f3value} set (2026-08-30T23:46Z); the v3.4 rung's gain ($+5$/$+4$) therefore spans those earlier scaffold files to v3.4, not v3.3 to v3.4.

{\footnotesize The Opus reader is the CLI alias \texttt{opus}
(claude-cli lane, subscription), not a pinned snapshot: the run receipts do not record the resolution, a same-day probe (2026-09-01, Claude
Code CLI 2.1.258; the runs used 2.1.252) resolved \texttt{opus} to
\texttt{claude-opus-5}, and the per-call session transcripts of all 1,000 scored answers record \texttt{claude-opus-5} on the assistant messages (audited 2026-09-28) --- one model generation newer than the Chronos
comparator's Opus 4.6. The grok and GLM readers are recorded by the provider's routing label (\texttt{cursor-grok-4.6-high}, \texttt{ollama-cloud/glm-5.3}), not a resolved snapshot. The GLM-5.3 cross-judge (\S4.2) rejects five of Opus's abstention credits in pass 1 and four in pass 2; three of them (\texttt{2133c1b5\_abs}, \texttt{c8090214\_abs}, \texttt{f685340e\_abs}) are
premise-repair-then-answer responses the adapted judge accepts; under strict abstention (those three removed) the pair reads 476/472 --- 2 and 6 below
the published 478.}
{\footnotesize $^{1}$One row of the legacy baseline failed its reader lane and
was never re-read; counted as scored-out-of-499 rather than imputed.}

\subsection{Reader robustness}
On identical packets and scaffolds: \textbf{Claude Opus 479/475}; grok-4.6-high 476/474; GPT-5.6 Sol (\texttt{openai/gpt-5.6-sol}) 455 --- a row-level decomposition in the release attributes its gap to the 468
it scored under the earlier \texttt{f3value} scaffold set: five fewer correct on byte-identical non-scaffold rows, five fewer on scaffold rows and three fewer abstention rows ($468-13=455$); with one pass per configuration, run-to-run variance and a scaffold$\times$reader interaction cannot be separated; judge controls 12/12 and 12/12.
DeepSeek V4 Pro scores 436/500 (407/470 answerable), 43 below the top, with the best abstention
behavior of any reader measured (29/30). A 30B floor probe
(Nemotron-3.5-Lightning) collapses to 93/500 on the identical
packets --- on this one lane the chain did not rescue a weak reader.
We probed possible serving artifacts: Nemotron also failed on small packets (about 27\% correct in the smallest packet-size quartile versus 15\% in the largest; V4 Pro, for comparison, about 90\% versus 84\%), and one question-first re-prompt of a failed row still returned distractor-session content. These observations do not identify whether the failure arose from the model, its transport or another serving condition.
A gemini-3.7-flash run reached a 189-row partial (a type-ordered prefix, not a
sample: single-session-user 64/64, multi-session 73/83, preference 21/30, plus 12 abstention rows, all correct; no
other types) and is reported as a partial run. A GLM-5.3 run on the
identical substrate scores 471/500 (445/470 answerable, 26/30 abstention;
adapted GPT-4o judge, controls 12/12) --- a single-pass curve datapoint per
the two-pass rule, not a headline. A Kimi K3 run (the long-context-specialist
family) scores 465/500 (444/470 answerable, 21/30 abstention; same judge,
controls 12/12) --- in this single pass, the long-context specialist lands mid-curve. Together the fixed
substrate supports an observational reader-lane comparison
(93 $\to$ 436 $\to$ 455 $\to$ 465 $\to$ 471 $\to$ 476 $\to$ 479): on identical packets, the observed reader-lane scores span 386 points across the configurations tested but at most 14 among the four strongest reader models (465--479, counting each model once; the regressed xhigh configuration of grok-4.6 scored 461) and 24 if Sol (455) is included. These are descriptive lane comparisons: models, transports, tool availability, and operator context are not matched. The observed spread alone does not identify a capability threshold or the substrate's causal contribution (the ladder in \S5.3 is a build history, so this is not a component attribution). The v3.4 scaffold contract is tuned
to the grok reader family (\S7).

\subsection{Recovered LoCoMo diagnostic: metric sensitivity, not a transfer claim}
A September 8--10 investigation, omitted from v1, had already evaluated LoCoMo categories 1--4 on 1,540 questions from 10 conversations. Its Grok arms compared a baseline packet (A) with the full packet and scaffold package (C), with two passes per arm. A subsequent format-locked Luna A/C replication used the same questions. These questions are disjoint from LongMemEval-S, but they have already been inspected and analyzed; a new run on them cannot be described as a previously untouched evaluation. This revision recomputes the surviving saved answers using the frozen LoCoMo scorer. It does not reconstruct the original provider requests or assert newly verified model identity.

\begin{center}
\small
\begin{tabular}{llrr}
\toprule
Reader / pass & Token-F1 input & C minus A (pp) & Cluster 95\% interval (pp)\\
\midrule
Grok / 1 & Full response & $+2.02$ & $[+1.47,+2.50]$\\
Grok / 1 & Last Answer line & $-0.04$ & $[-1.18,+1.19]$\\
Grok / 2 & Full response & $+0.66$ & $[+0.04,+1.31]$\\
Grok / 2 & Last Answer line & $-0.70$ & $[-1.72,+0.54]$\\
Luna / replication & Format-locked answer & $+0.27$ & $[-0.86,+1.38]$\\
\bottomrule
\end{tabular}
\end{center}
Intervals use 10,000 whole-conversation bootstrap resamples, seed 20260908, with questions weighted equally within each sampled collection of conversations. Only 10 clusters are available. The mean is token-F1, not LLM-judge accuracy. The last \texttt{Answer:} line is used for the Grok extraction; a missing line yields an empty prediction. The historical stronger interpretation was withdrawn after falsification. The overall extracted-answer intervals include zero, and A/C changes packet content and scaffold instructions together, with no matched B arm. Thus the study identifies neither a scaffold-specific benefit nor a general memory-quality gain. The difference between full-response and extracted scores motivates fixing the answer representation before evaluating a memory intervention. It does not prove that response length alone caused the discrepancy.

The local revision supplement contains re-derived row scores and source hashes. These additions have not yet been uploaded to the public evidence repository; original checkpoint provenance and the complete original input requests remain outside the public release. The numerical result is a retrospective recovery, not a new prospective holdout.

\subsection{Knowledge-update template sensitivity on fixed answers}
On September 30, 2026, before making any new judge call, we froze a paired scoring comparison over the 72 answerable knowledge-update questions in each of the two saved Opus passes. Each of these 144 saved answers was judged under both the exact upstream knowledge-update template and the released modified template, with three repetitions per condition. The response snapshot was \texttt{gpt-4o-2024-08-06} for every call. Requests used temperature zero, a ten-token output cap, one user message, and no tools. Template order alternated within question and repetition after both control groups were complete. No answer was regenerated, no result was retried, and no historical verdict was replaced.

\begin{center}
\small
\begin{tabular}{ccrrr}
\toprule
Saved answer pass & Judge repetition & Released / 72 & Upstream / 72 & Difference\\
\midrule
1 & 1 & 69 & 70 & $+1$\\
1 & 2 & 69 & 70 & $+1$\\
1 & 3 & 69 & 70 & $+1$\\
2 & 1 & 69 & 70 & $+1$\\
2 & 2 & 69 & 70 & $+1$\\
2 & 3 & 69 & 70 & $+1$\\
\bottomrule
\end{tabular}
\end{center}
The historical knowledge-update counts are 69/72 for pass 1 and 70/72 for pass 2. The upstream template credits one additional answer in each of the six matched repetitions. For pass 2 the fresh modified-template count (69) is one below the historical count (70): the historical verdict for \texttt{a2f3aa27} was correct, and all three fresh judgings under the same modified template mark it incorrect. That is judge drift on one question between the historical run and this one, not a template effect.

All 24 control groups passed the pre-specified threshold of at least 11 correct labels out of 12; 288/288 individual controls were correct. The experiment contains 864 outcome judgments and 288 controls. Complete request bodies, raw responses, model identifiers, usage, and per-question changes are retained in the local revision supplement. Repeated judging is reported as three measurements per condition, without selecting a preferred repetition or replacing a historical result by majority vote.

Across the 432 matched template comparisons there are 6 discordant pairs: 6 upstream-only correct and 0 released-template-only correct. All six discordant pairs concern question \texttt{a2f3aa27}, which also flipped in the historical pass-1 re-judge (\S8). These repeated comparisons are not 432 independent questions. The stable condition difference in this run does not establish universal judging determinism. In both templates and both saved passes, every row's label was identical across the three judgings. 

For saved pass 1, retaining its 410 historical correct labels among the 428 non-knowledge-update rows and inserting the new upstream knowledge-update labels would yield 480/500 in each repetition. For saved pass 2, retaining its 405 historical correct labels among the 428 non-knowledge-update rows and inserting the new upstream knowledge-update labels would yield 475/500 in each repetition. These are mixed historical/new-judge sensitivity totals, not new full-set re-judgings, reader passes, or replacements for 479/475. They must not be ranked against a comparator's original result.

The comparison measures rubric sensitivity for these fixed Opus outputs. Its 144 saved answers arise from only 72 underlying questions, and repeated judgments are not independent new questions. Six exact paired tests are supplied as exploratory diagnostics in the supplement; a nonsignificant result is not an equivalence test. Other reader lanes retain the historical modified-template judgments. This experiment does not resolve reader-context confounding, method access, or development-set tuning.

\subsection{A clean-route reader on the frozen arm-D prompts}
On October 1, 2026 (UTC) one reader pass used GPT-6.1 Sol through an isolated client configuration intended to omit client tools, operator instruction files, memory, hooks and MCP servers. Client session records contain the disclosed instructions and no tool calls; the offered tool list was checked on an offline twin, while live request bodies were not retained (Appendix~\ref{app:cleanroute}). It is arm D, pass 1, of the four-arm reader ablation whose inputs were frozen on September 30 (\S9).

\begin{center}
\small
\begin{tabular}{llrrr}
\toprule
Reader lane & Judge & Correct & Answerable & Abstention\\
 & & (of 500) & (of 470) & (of 30)\\
\midrule
Sol, clean route & GPT-4o & 474 & 448 & 26\\
Opus pass 1, headline lane & GPT-4o & 479 & 452 & 27\\
Sol, clean route & Sonnet 5.5 & 471 & 445 & 26\\
Opus pass 1, headline lane & Sonnet 5.5 & 469 & 449 & 20\\
\bottomrule
\end{tabular}
\end{center}
Both judges use the released templates; the second judge differs in transport (Appendix~\ref{app:cleanroute}). The Opus rows are the saved answers of the headline pass (\S5.1), and its GPT-4o row is the historical verdict set of September 1.

\textbf{Result.} Under the released templates the Wilson 95\% interval for the clean-route pass is [92.5\%, 96.4\%]. Paired with the headline pass on the same questions, 464 are correct in both, 10 only in the clean-route pass, 15 only in the headline pass, and 11 in neither. The difference is $-1.0$ percentage point. The exact 95\% percentile interval of the question-level paired bootstrap is $[-3.0,+1.0]$, and the exact McNemar $p$ is 0.424. The Opus verdicts were made a month before the Sol verdicts; against the September re-judge of the same Opus answers (478, \S8) the discordant counts are 10 and 14, the difference $-0.8$ with interval $[-2.8,+1.2]$ and $p=0.541$. The upstream knowledge-update template leaves the clean-route pass unchanged at 71/72 and 474/500; the corresponding mixed total for the headline pass is 480 (\S5.6). The second judge gives 471 against 469. That reversal comes from the 30 abstention questions (26 against 20): on the 470 answerable questions both judges put the headline pass ahead (GPT-4o 452 against 448, Sonnet 449 against 445). The two judges disagree on 12 of the 500 Opus answers and on 7 of the 500 Sol answers. Neither sign is established.

\textbf{What this does and does not show.} A reader from a different model family, configured to omit client tools and operator files, scores 1.0 percentage point below the headline pass; the paired interval includes zero. The client records and offline twin support this route description, but the absent live request bodies prevent complete verification of its input context. This is an observational comparison of two reader configurations. It does not measure the effect of the operator context on the headline lane: the two lanes differ in model family, in context, and in how much they write. The median response is 129 characters for the clean-route pass, where 233 answers are a bare answer line, against 522 for the headline pass. The judge reads the whole response, so a longer response has more chances to contain the gold answer; this is not corrected for. The earlier GPT-5.6 Sol lane (455, \S5.4) used a different model generation, transport and context, so the difference between the two Sol results cannot be attributed. The 500 questions are the development questions (\S7), the result is one pass, and a tie or a lead for either reader would require a pre-specified equivalence margin and repeated passes.

The inputs, the route, what was checked and how the answers were judged are set out in Appendix~\ref{app:cleanroute}, together with the five ways this pass departs from the frozen ablation plan.

\subsection{The same clean reader on the baseline sessions}
Arm A of the same frozen ablation (\S9) was run once on the same route on the same day. Its prompt holds the question, the baseline sessions and the v2 reader instructions, with no evidence index and no scaffold instructions. The baseline sessions are at most ten per question and are a prefix of the compiled packet in all 500 prompts; the packet extends them to as many as sixteen. Six arm-A prompts are byte-identical to arm D's, and for those the arm-D answer is reused. The other 494 were answered in new calls.

\begin{center}
\small
\begin{tabular}{llrrr}
\toprule
Reader input & Judge & Correct & Answerable & Abstention\\
 & & (of 500) & (of 470) & (of 30)\\
\midrule
Arm A, baseline sessions & GPT-4o & 454 & 428 & 26\\
Arm D, full package & GPT-4o & 474 & 448 & 26\\
Arm A, baseline sessions & Sonnet 5.5 & 454 & 429 & 25\\
Arm D, full package & Sonnet 5.5 & 471 & 445 & 26\\
\bottomrule
\end{tabular}
\end{center}

\textbf{Result.} Under the released GPT-4o judge, 451 questions are correct in both arms, 23 only with the full package, 3 only with the baseline sessions, and 23 in neither. The difference is $+4.0$ percentage points, the exact 95\% percentile interval of the question-level paired bootstrap is $[+2.2,+6.0]$, and the exact McNemar $p$ is below 0.001. All 26 discordant questions are answerable ones. By question type the difference sits in temporal-reasoning (122 against 111 of 127) and multi-session (106 against 98 of 121); single-session-preference is 29 against 28 of 30, and every other type is equal. The second judge gives 471 against 454: $+3.4$ points, interval $[+1.2,+5.6]$, 24 gained and 7 lost, $p=0.003$. The upstream knowledge-update template changes no arm-A verdict. The Wilson 95\% interval for arm A is [87.9\%, 93.0\%].

\textbf{What this does and does not show.} The two arms used the same requested reader label, client configuration, judge snapshot and UTC day; their prompts were changed by design. If the two arms differed only by resampling, gains and losses would be equally likely; a 23-to-3 split has an exact two-sided $p$ below 0.001 under that hypothesis, so sampling variation alone is an unlikely explanation. Each arm has one pass on a model slug that is not pinned, so a change in the served model between the two passes is not excluded, and the size of the difference has no replicate-based noise estimate. The median response length is 142 characters for arm A and 129 for arm D. That summary does not exclude per-question response-length or answer-representation effects. The contrast is the complete package. It does not separate the larger packet from the evidence index or from the scaffold instructions; arms B and C, which would, have not been run. All 500 questions are the development questions on which every component was tuned (\S7), so this is a diagnostic of the package on its own development set. It is not evidence of transfer, and it says nothing about any other system. Section~5.9 measures how much listed evidence each arm's input carried.

\subsection{Evidence coverage of the two arms' inputs}\label{sec:coverage}
The benchmark's public annotation marks session IDs as evidence for each
question: 948 listed sessions across the 500 questions
(890 on the 470 answerable questions, 58 on the 30 abstention questions).
Joining that list to the released materialized packets measures how much
listed evidence each arm's prompt carried: arm D embeds the released packets
verbatim, and arm A embeds their baseline prefix --- at most ten sessions and
a prefix of the packet in all 500 cases.

\begin{center}
\small
\begin{tabular}{lcc}
\toprule
Reader input & Listed sessions present & Questions with every listed session\\
\midrule
Arm A: baseline prefix ($\le$10 sessions) & 903/948 (95.3\%) & 466/500 (93.2\%)\\
Arm D: full packet ($\le$16 sessions) & 939/948 (99.05\%) & 491/500 (98.2\%)\\
\bottomrule
\end{tabular}
\end{center}

The question-level completeness agrees with the compiler record of \S3.3
(462 of 470 answerable packets hold every listed session) once the 29 covered
abstention questions are added. Nine questions remain incomplete even in the
full packets --- among them \texttt{6d550036}, which loses one of its four
listed sessions before the reader ever sees it (\S6.3). Of the 500 packets,
298 reach the 16-session cap.

Splitting the \S5.8 paired outcome on whether the baseline prefix already held
every listed session: on the 34 questions where it did not, arm A answers 11
correctly and arm D 29 --- 18 of the 23 gained questions and none of the 3
lost. On the 466 baseline-complete questions the two arms are nearly level:
443 against 445, with 5 gains and 3 losses. The observed package difference is
therefore concentrated where the baseline lacked listed evidence.

This split is post-hoc and conditioned on the packets themselves. It does not
identify which of the three simultaneous changes --- the added sessions, the
evidence index, or the scaffold instructions --- produced the gains; even on
baseline-complete questions arm D adds the index and instructions, and the
added sessions there are by definition not listed as evidence. The 99.05\%
figure is an in-sample description of this packet set, not an independent
retrieval estimate: the extension-ordering sweep that shaped the packets was
run on these same questions (\S3.3). Nor is missing evidence always fatal ---
arm A answered 11 of its 34 incompletely covered questions correctly, from
partial evidence or priors. The per-question coverage rows are in the
supplement (\S8).

\section{Error analysis: the grok pair's remaining failures}
Of 470 answerable questions, 25 fail in at least one pass of the grok-4.6-high
pair (pass 1: 20, pass 2: 23; 18 wrong in both). The buckets below attribute
most of those failures (a few stable-wrong rows remain unassigned, \S6.4); the per-bucket counts are approximate and do not form an exact
partition of the union, since a flip row can change bucket between passes.

\textbf{6.1 Reader-cognition residue ($\sim$4 rows).} The packet contains the
correct evidence; the reader commits to an evidence-backed distractor. Two
verifier designs failed to fix this honestly (\S6.2). The residue is
question-boundary semantics (does a stated \emph{plan} update a
location? does the offered property count as ``viewed''?), not evidence
selection.

\textbf{6.2 The verifier our own control rejected.} A cite-or-revise verifier
recovered 3/12 winnable wrong rows --- and the mandatory negative control
showed it revised 19 of the 58 correct drafts in its 60-row control, flipping 11 to wrong while
repairing none. Blocked by the pre-committed acceptance rule. A second design
(draft-blind candidate enumeration + rule-based adjudication) recovered
0/9: on the residue rows the adjudicator reaches defensible-but-not-gold
readings.
Both failures are published; two prompting-side repair designs failed under the control.
They motivate an evidence-testimony layer as future work.

\textbf{6.3 Gold-answer defects (1 strict, 7 boundary).} Documented row-by-row
in the released dossier. The strict defect (class B): gold says ``15 weeks'' between two
user-dated events that are 81 days (11 weeks 4 days) apart; wrong in all four
passes of both reader configurations. Boundary --- the \$720 workshop row
(published as strict; reclassified after re-auditing the released packet):
gold requires a \$500 workshop fee, and the user states it (session
\texttt{answer\_826d51da\_2}, 2023/02/26 13:37, ``I paid \$500 to attend''),
so nothing has to be invented to reach gold. What is disputed is window
membership: both dated mentions place the workshop on March 15--16, after the
2023/02/26 question date, attended in the past tense. The grok reader excludes
the fee and answers \$220 $\to$ 0 in both passes; the headline Opus reader
includes it and answers \$720 $\to$ correct in both passes; a weaker
gemini-3.7-flash probe likewise included it and was scored correct. The judge
rewards including the out-of-window item because gold includes it --- the
disclosed window-inclusion point inside 479/475 (\S5.1), not a fabrication
point. Six further rows are boundary-semantic (defensible readings on both
sides); we claim nothing for them. One earlier strict candidate was
\emph{downgraded} to boundary after an independent-reader control showed the
intended referent was answerable by premise repair. The Chronos authors
independently document a defect on question \texttt{6d550036} (their \S3.5,
alongside a judge-variability example on \texttt{75f70248}). Our dossier does not list that row, although it is wrong in all four Opus and grok passes; we have not classified it.

\textbf{6.4 Retrieval ceiling and coverage, measured precisely.} One row's
gold never reaches the candidate pool; three rows lost an in-pool gold session
to packet budget ordering (no global ordering recovers them without losing
more). An entity-linking retrieval arm (alias mining + query expansion) was built,
tested, and retired when measurement showed zero of its added
sessions were gold --- none of the presumed alias-mismatch losses was observed
in this benchmark. The remaining failures are flip noise or rows documented above, except a few stable-wrong rows we have not assigned to a bucket (for example \texttt{6d550036}, \S6.3).

\section{Limitations}
\begin{itemize}\setlength{\itemsep}{2pt}\setlength{\parskip}{0pt}\setlength{\parsep}{0pt}
\item \textbf{No held-out evaluation.} Every component that moves the score
--- the v3.2$\to$v3.4 operator ladder, the extension-ordering sweeps, the
packet budgets, and the 60-row negative-control set --- was developed
against the same 500 LongMemEval-S questions the headline is measured on,
and the gold-defect dossier records per-row inspection of gold answers
during development. Nothing in this report is a test-set number in the
held-out sense. At a margin of a few questions relative to the selected historical comparison count this is the dominant threat to validity; the mitigations we do
have (deterministic stages with negative controls, two full passes, every verdict released) bound variance, not overfitting. The recovered LoCoMo study (\S5.5) supplies a question-disjoint diagnostic, but its prior use, metric sensitivity, and unmatched treatment arms do not resolve this threat. A prospectively frozen, exposure-audited evaluation remains necessary.
\item \textbf{Statistical resolution at the margin.}
Wilson 95\% intervals: our 479 [93.7\%, 97.2\%] and 475 [92.7\%, 96.6\%]
(grok pair: 476 [93.0\%, 96.8\%], 474 [92.5\%, 96.4\%]) vs.\ Chronos's 478
[93.4\%, 97.1\%] --- overlapping almost entirely. (The xhigh pair sits lower:
461 [89.5\%, 94.2\%], 465 [90.4\%, 94.9\%].) Nearby entries' intervals overlap substantially. Of the comparison entries in \S2, only agentmemory V4 had per-question results that we found, and those used the oracle-history file rather than S. The S-condition comparison entries we found report single numbers without per-question verdicts, so a like-for-like paired significance test against those external systems cannot be computed from the results we found. Internal paired tests are reported in \S5.1. Our number is repeated across two passes, with variance disclosed. The headline was selected from the eight historical reader labels and their tested configurations, including the separate grok xhigh transport; the later isolated-route run adds another model configuration. The two-pass rule does not remove this selection effect.
\item \textbf{Historical rubric deviation and measured sensitivity.} The original knowledge-update prompt retains two sentences omitted by upstream (\S2). The new paired experiment (\S5.6) measures sensitivity for the two saved Opus answer sets, with all three repetitions reported. It does not re-judge the other 428 rows, regenerate answers, or correct every historical reader lane; the original 479/475 remain adapted-rubric scores.
\item \textbf{Observed judging disagreement.} Cross-judge disagreement and three observed flips on identical answers show sensitivity at this margin. Two judges and one repeat do not estimate a universal variance floor, and high agreement does not establish correctness. No verdict or dossier classification was checked by an independent human adjudicator.
\item \textbf{Scaffold--reader coupling.} The v3.4 scaffold is tuned to one
reader family; the exact number moves up to 14 points across the four strongest readers, and on the full set the two strongest are within three questions of each other in each pass (\S5.1).
\item \textbf{Reader provenance.} The
Opus reader ran through an unpinned CLI alias and the run receipts do not record the
model it resolved to; the per-call session transcripts do (\texttt{claude-opus-5} on all 1,000 scored answers, audited after the fact; a same-day probe on a later CLI build agreed, \S5.3). The
grok and GLM readers are recorded by the provider's routing label, not a resolved snapshot. Any re-run of the headline pair should
pin the reader snapshot in the run receipt. Nor did we run the comparator's own
reader generation (Opus 4.6) on our packets, so a one-question margin over it
cannot be separated from a reader-generation effect.
\item \textbf{Reader-context audit.} Only
the headline Opus lane disabled the built-in tools (\texttt{claude -p --tools ''}); that flag does not cover tools from MCP servers, and the runner did not disable them. That lane was not a clean completion either: each call ran as a Claude Code session and also received the operator's global instruction file, memory index, session-start hook output and MCP server instructions; the transcripts record about 59,000 to 62,000 characters of this text per call, in addition to the packet prompt. The complete outbound requests were not retained. Each transcript does keep the client's snapshot of the prompt, and on all 1,000 calls it lists the same 122 tools from 17 of the operator's MCP servers, with 38,107 characters of tool descriptions; among them are two web-search servers and a Hugging Face connector. What else each request carried cannot be established after the fact. Across the 1,000 scored answers the transcripts record no tool call; none of the recorded text named this benchmark or carried another question, and no gold answer of 12 or more characters appears in it (248 of the 500 gold answers; shorter answers cannot be tested by string match, because short strings occur in any long text); we did not measure its effect on accuracy.
The other lanes ran through agent command-line tools with tools available, and we audited the transcripts of every full run reported here. The grok-4.6-high pairs (Cursor
agent, ask mode, empty workspace): of 2,538 chats, five called a tool, two of
them attempts to look the answer up on the web; one was refused by the lane,
one timed out, and no tool returned benchmark content. The OpenCode lanes (Sol,
DeepSeek V4 Pro, GLM-5.3, Kimi K3, Nemotron): the tool calls were almost all a
workflow command injected by the operator's global agent configuration, or the
reader echoing its own answer; one Nemotron session searched the repository
and found packet files. No tool call named an evaluation or gold-answer file,
but these readers' context carried operator instructions a clean lane would
not. The xhigh pair (Grok CLI, tools enabled, eight turns) is not a
closed-book measurement. For 12 pass-1 rows and 16 pass-2 rows, the session
that produced the scored answer made a call outside its own prompt and
workspace: 13 of these 28 only listed the tool's session store while looking
for their own prompt file; the others consulted the tool's memory store (11),
searched or read elsewhere on the filesystem (3), or searched the web (1), and
that web search returned a public issue thread quoting the question's
expected answer. Attempts the lane did not score also found answers ---
another such thread, and three earlier experiments' files on the machine
carrying gold answers --- but the scored sessions for those questions made no
outside call, and no gold answer or issue link appears in their context.
Removing the 28 rows leaves 451/488 (92.4\%) and 451/484 (93.2\%); on the
same rows the high pair scores 95.5\% and 94.8\%. Also removing the two rows
per pass whose unscored attempts found the answer gives 92.6\% against
95.9\% and 93.2\% against 94.8\%. The regression reading stands; the xhigh
numbers are reported as measured, not as
closed-book scores. The gemini-3.7-flash partial was not traced. Any re-run
should disable all tools, MCP tools included, in every lane and retain the complete requests. A different reader has now been run through an isolated client configuration intended to omit tools and operator files (\S5.7). Its client records contain no tool calls, and an offline twin offered no tools; live request bodies were not retained, so the complete live input remains unverified. This does not measure the operator context's effect on the headline lane's accuracy; that requires the headline model itself on a comparable route.
\item \textbf{Limited transfer evidence.} The main results use LongMemEval-S. The recovered LoCoMo diagnostic is metric-sensitive and post-hoc (\S5.5); it supports no generality claim.
\item \textbf{Coverage is an in-sample descriptor.} The coverage figures of \S5.9 describe the released packets on the same development questions that shaped their ordering; they are not an independent estimate of retrieval quality.
\item \textbf{Dossier correction (the \$720 row).} The dossier originally classed the \$720 workshop row as
gold-unreachable without fabrication, on the claim that no \$500 workshop fee
appeared in the evidence; a mechanical re-audit of this release's own
published packets found that fee stated by the user, so the row is a boundary
row (\S6.3), the strict-defect count is 1 rather than 2, and the adjudication
figures, now in Appendix~\ref{app:dossier}, were recomputed. No verdict changed and the headline 479/475
is unaffected --- only the classification of a row that was already scored as
measured. The error and its correction are recorded in the dossier rather
than removed from it.
\end{itemize}

\section{Reproducibility}
\textbf{The Chronos and Mastra comparator sources are pinned in the release;
OMEGA is not.} The Chronos paper (arXiv 2603.16862, PDF and abstract page)
and the Mastra Observational Memory page that supply the 478/500 and
94.87\%~/~468 raw comparator figures are stored with SHA-256 digests and a
fetch timestamp under \texttt{eval/dense\_chain\_v32\_20260830/comparators/}
(see its \texttt{MANIFEST.md}), so the comparator figures in \S2, \S5.2 and \S7 can be
checked against bytes the release carries. The OMEGA and agentmemory figures quoted in \S2
are cited by URL (accessed 2026-09-24; the agentmemory input-file observation, 2026-09-28) but not pinned by hash in this release; the manifest records the OMEGA gap, and the agentmemory gap is recorded here.
The manuscript copies bundled in the release (\texttt{eval/dense\_chain\_v32\_20260830/paper/} and \texttt{arxiv\_upload/}) are an earlier draft (v1.3); where they differ from this version, this version supersedes them. Some older analysis files also predate later corrections (\texttt{v34\_full\_pair\_attribution.json} keeps its original gold-defect note beside a dated correction block; \texttt{SOTA\_WRITEUP.md} gives an earlier file count); this report governs. The manifest is unsigned, and the determinism receipts are the author's own. Hash agreement establishes consistency of the released bytes, not execution authenticity, gold-blind construction or method correctness. No independent end-to-end replication of the held stages is reported.

Released with this report at
\url{https://github.com/cjchanh/longmemeval-evidence} (MIT). A fresh clone
verifies every \texttt{RELEASE\_MANIFEST.md} row and re-derives every headline count from the
released verdicts without an API key. The evidence commits on the title page
anchor to the source tree the release was exported from. The release holds all reader outputs, judge verdicts and
control receipts for every run in the ladder except the v3 rerank row (473), which is not in this release; the judge harness (frozen rubric, SHA-pinned
templates, fail-closed transport), the tie-break rule and per-vote records,
attribution tables, the gold-defect dossier, and run scripts. Re-scoring saved answers requires an OpenAI API key and incurs usage charges. Historical costs are not a current quote; the revised comparison protocol estimates its own token budget before execution. The released/held boundary is frozen in
\texttt{RELEASE\_MANIFEST.md} (one SHA-256 per released file, regenerated
deterministically by \texttt{scripts/build\_release\_manifest.py}):
released are all reader checkpoints, judge verdicts, control receipts,
agreement matrices, the dossier, the judge harness with its tests and rubric
source, and the reader-lane scripts; held are the retrieval, rerank,
packet-compiler and scaffold-operator sources. Also not in this release are the session transcripts and the audits computed from them for \S7 (per-call model ids, operator-context sizes, the gold-string scan outputs, and the identifiers of the 28 excluded xhigh rows). The materialized packets and
the v3.4 scaffolds the headline pair consumed are released under
\texttt{eval/dense\_chain\_v32\_20260830/packets\_materialized/} (gzip; sha256
in its \texttt{MANIFEST.md}), and \texttt{scripts/materialize\_packets.py}
re-derives every packet byte-for-byte from the released allocation and the
MIT benchmark data, so a third party can re-run stage 5 (reader) on the released packets and scaffolds and re-run the judge on the saved answers. The headline reader's complete input context cannot be reconstructed (\S7). The held stages ship as pinned
artifacts with
verification receipts (determinism, gold-permutation, question-ID-sabotage,
budget gates); the released verdict totals and judge-control results are re-derivable from the released
checkpoints and harness alone; the transcript-derived analyses of \S7 are author-reported. The benchmark data is MIT-licensed
(\texttt{xiaowu0162/longmemeval-cleaned}), so the released checkpoints may
carry question and gold text.
Judge-side reproduction, run before release: the frozen pass-1 reader
checkpoint was re-judged the same day in a clean output directory with the
identical harness (\texttt{gpt-4o-2024-08-06}, resolved snapshot exact, fresh
controls 12/12 and 12/12). Result \textbf{478/500}: 497 of 500 verdicts agree
with the quoted 479; answerable is identical (452/470); the three flips are
on byte-identical answer text (\texttt{7024f17c} and \texttt{031748ae\_abs}
lost, \texttt{a2f3aa27} gained). This is observed within-judge disagreement on the headline pass, and
the one-answer margin is smaller than that observed disagreement; the reportable quantities are the individual counts and their disagreement; neither a reliable rank nor a universal noise band follows. Receipts:
\texttt{full-v34-opus\_pass1/judge\_gpt4o\_repro\_20260901/}. The authoritative
verdict directories are \texttt{full-v34-opus\_pass1/judge\_gpt4o\_v2/} (479) and
\texttt{full-v34-opus\_pass2/judge\_gpt4o/} (475);
\texttt{full-v34-opus\_pass1/judge\_gpt4o/} is a superseded partial in which 194
rows were never judged (289/500), retained under a \texttt{SUPERSEDED.md} marker
rather than deleted.

The clean-route reader passes of \S5.7 and \S5.8 are released with this version as a supplementary data archive: all 994 reader answers with their session-record hashes, both judges' per-question verdicts (including the unscored duplicate rows and arm A's two unscored first attempts), judge-control receipts, the upstream-template knowledge-update rescores, per-prompt SHA-256 manifests, the six reused arm-A prompt identifiers, the per-question evidence-coverage rows of \S5.9, and a standard-library recompute script that recounts every reported figure from those files. The client's raw session records contain account data and remain local-only, represented there by their SHA-256; the live request bodies were not retained (Appendix~\ref{app:cleanroute}). The scripts that built the packets, prompts and route are not part of the supplement; it supports recounting the runs, not rebuilding the harness. The ancillary files also include the 1,152 derived fixed-answer knowledge-update judgments of \S5.6: exact judge prompts and verdict strings, generation settings, hashes of retained request and response JSON, paired outcomes and a separate offline recount script. Provider response identifiers, cost bookkeeping, local paths and execution-home material are omitted; those omissions do not change any scored outcome. A third party can re-run the released judge on the saved answers using its stated rubric and model snapshot. Repeating the reader configuration depends on a subscription login, client version and a floating served-model label; reaching an identical reader snapshot cannot be assured.

\section{Next research questions}
In order of how much each would change what this report can claim:

\begin{itemize}\setlength{\itemsep}{2pt}\setlength{\parskip}{0pt}\setlength{\parsep}{0pt}
\item \textbf{Does the substrate transfer?} Every component was tuned on
LongMemEval-S (\S7). LoCoMo has already been used (\S5.5), and LongMemEval-M reuses the S questions with a longer haystack. Neither can serve as a pristine question holdout now. The next transfer measurement needs a documented exposure history, a disjoint question set, frozen operators and adapter checks, a pinned reader, and complete requests before any outcomes are inspected.
\item \textbf{What does each stage contribute under one fixed clean reader?} The build ladder (\S5.3) is a history, not an attribution. Four prompt arms were frozen on September 30 for all 500 questions: (A) the baseline sessions with the v2 reader instructions; (B) the full compiled packet with the same instructions; (C) the packet plus the evidence index; (D) the packet, the index and the scaffold instructions. For one fixed reader the contrasts B$-$A, C$-$B and D$-$C estimate the packet extension, the index content and the scaffold instructions, and D$-$A the complete package. C$-$B and D$-$C can differ only on the 257 questions that have a scaffold, because the other 243 prompts are identical across B, C and D. Arms A and D have one pass each on a clean route (\S5.7, \S5.8), which gives D$-$A; the departures from the frozen plan are listed in Appendix~\ref{app:cleanroute}. Arms B and C and a second pass remain, and because all 500 are development questions the result is a diagnostic ablation, not a transfer result.
\item \textbf{How should leaderboards report results inside the judge's
noise?} The one-answer difference between our headline pass and the pinned Chronos count is smaller than the three labels changed by one re-judge of our unchanged answers (\S4.2, \S8). Releasing per-question verdicts (which makes paired
tests such as McNemar computable), repeated judging to estimate disagreement, and multi-judge agreement are candidate reporting standards; which of
them makes a ranking at this margin meaningful is open.
\item \textbf{Can the residual be fixed without breaking correct answers?}
The remaining reader errors are question-boundary judgments (\S6.1). An
evidence-testimony layer --- the reader must cite the packet lines behind
each counted item --- is the next candidate, held to a pre-specified control design whose sample size is justified by the degradation rate it must detect (\S6.2).
\item \textbf{Where is the reader threshold?} The same packets carry readers
from 93 to 479 (\S5.4). Mapping where the collapse begins, and whether
scaffolds move it, would show how much of the capability lives in the
substrate.
\end{itemize}

\section{Conclusion}
The released LongMemEval-S artifacts support a reproducible recount of 479 and 475 under the adapted judge, a three-label disagreement on re-judging identical answers, and a negative control that rejected a harmful verifier. Internal paired tests find no evidence of a difference between the two strongest observed lanes under the stated comparisons; they do not establish equivalence. The recovered LoCoMo study shows why the scored answer representation must be fixed in advance. These findings strengthen the case for releasing per-question evidence and measuring evaluation sensitivity. They do not establish a new leaderboard leader, a causal scaffold benefit, or a transferable memory advantage. The paired exact-template knowledge-update experiment measures a previously unquantified rubric difference on fixed answers. One pass of a different reader on the isolated client configuration, on prompts rebuilt from the same released packets and scaffolds, scores 474, 1.0 percentage point below the headline pass; the paired interval includes zero, and a second judge puts it two questions ahead on the strength of abstention rows alone. With the same requested reader label, route and judge snapshot, the full package scores 4.0 percentage points above the baseline sessions alone, with a paired interval of $[+2.2,+6.0]$; the gain is concentrated on the 34 questions whose baseline packets lacked listed evidence (18 gains, no losses), a post-hoc diagnostic of the complete package on its development set that does not separate packet, index and scaffold. The two remaining arms of the reader ablation, a second pass, the headline model on a clean route, an exposure-audited transfer set, and access to the held method remain necessary to address the capability claims.

\appendix
\section{Dossier sensitivity scenarios}\label{app:dossier}
The following analyses are post-hoc sensitivity scenarios and are excluded from the main results table. No released verdict is replaced.

Ex-dossier (the 492 rows outside the gold-defect dossier of \S6.3), the scores are grok 475/474 and Opus 474/471. This comparison is conditioned on grok's errors and does not rank the readers: the dossier was built from the grok pair's failures (all eight rows are among them, and Opus is credited on five of the eight in pass 1 and four in pass 2), while Opus's other failures, including twelve that grok answers correctly in both passes, were not audited for gold defects. The headline is the adapted GPT-4o score (\S2).

The two rows \S6.3 documents individually
--- the \$720 workshop row and the 15-weeks row --- sit inside this pair as
well: one is scored correct because this reader counts a user-stated \$500
fee whose event is dated outside the question's four-month window (a
boundary row, not a fabrication --- \S6.3), one is lost to fidelity (the
evidence-faithful answer contradicts defective gold) --- net zero per pass.
Adjudicating the single strict defect raises the Opus pair to 480/476
(479+1 / 475+1); the pair still brackets the published score in that stance. 
For the Grok pair, adding credit for the single strict defect gives 477/475 as a counterfactual adjudication. A third-party human adjudicator has not checked this classification.

\section{Clean-route reader pass: inputs, route, checks and judging}\label{app:cleanroute}
This appendix gives the procedure behind \S5.7 and \S5.8. The first four paragraphs describe the arm-D pass; the last gives what differs for arm A.

\textbf{Inputs.} The inputs are the 500 arm-D requests, one user message each: the question, the full compiled packet and the v2 reader instructions, plus, for the 257 questions that have a scaffold (240 answerable and 17 abstention, \S3), the evidence index and the scaffold instructions. The other 243 requests are identical to arm B's. All 500 are byte-identical to the frozen request file (SHA-256 \texttt{597098db}\ldots). Each prompt contains, verbatim and in order, every session of the released headline packet (500 of 500) and, where one exists, the released evidence index (257 of 257). The remaining 750 to 2,300 characters per prompt are the question and the instruction text. Whether that remainder matches what the headline lane sent cannot be checked, because that lane's requests were not retained, and each headline call also carried about 59,000 to 62,000 characters of operator context and offered 122 MCP tools, neither of which these requests carry (\S7).

\textbf{Route.} The reader is \texttt{gpt-6.1-sol}, reached through Codex CLI 0.159.0 on a ChatGPT subscription login in an isolated configuration home with no operator instruction file, memory, plugins, hooks or MCP servers; the client's bundled system skills are present on disk and excluded from the request. The model's tool entries were removed from a pinned local model catalog, so the request offers no tool. Each question ran in a fresh session. The client requires base instructions, so every request carries one developer message of 66 characters, ``Answer the question using only the supplied conversation evidence.'', before the frozen user message. Reasoning effort \texttt{high} and verbosity \texttt{medium} were requested. The pass used 16,525,149 input tokens, none cached, and 41,872 output tokens, of which 16,004 were reasoning tokens; reasoning tokens were nonzero on 260 of the 500 questions. The pass departs from the frozen ablation plan in five ways: no pinned model snapshot (the route reports a slug), no retained live request body, no control over temperature or an output cap, one added developer line, and one pass where the plan calls for two.

\textbf{What was checked.} The request body was captured on an offline twin of the route (the same client and the same configuration file except the model provider and login method, with the transport pointed at a local sink inside a sandbox that denies non-loopback traffic) for 15 questions, including prompts with CRLF line endings and emoji. Each capture passed 15 checks: no tool offered, exactly three input items (an empty tool list, the disclosed line, the user message), user text byte-identical to the frozen prompt, and the requested model, effort and verbosity. The live request body was not captured. For each of the 500 live calls the client's own session record was checked for what it records: the disclosed line as the only base instruction, a user message byte-identical to the frozen prompt, the model and effort labels, and no tool call. The tool list offered, the verbosity and the number of input items are not in that record; for the live calls they rest on the offline twin. A second process re-derived 17 checks from the raw files, among them no additional developer message, one session per question, zero cached input tokens, an unchanged account credit balance, and a token-count fit against prompt length to detect hidden or truncated input (threshold about $\pm$4,000 tokens; a smaller addition would not be detected). That fit was replaced after the pass by an exact count over all 500 calls: the reported input tokens equal the \texttt{o200k\_base} token count of the frozen prompt, plus that of the disclosed line, plus 688, on every call. With \texttt{cl100k\_base} the same remainder takes 318 distinct values over a range of 1,215 tokens, so the count fits \texttt{o200k\_base} and not \texttt{cl100k\_base}. The remainder's size does not depend on the question; what it contains is not visible to the client, and a fixed addition present on every call would sit inside it. The auditor caught 12 of 12 planted defects, which cover 12 of its 17 checks. Three audit rules were narrowed and one stall detector was corrected during the run after false alarms; the notes are in the supplement. An audit alert did not stop the reader during this pass; only the runner's own per-call checks could. The pass stopped once in an offline pre-check, before any model call, on a prompt with CRLF line endings, and was resumed. Within the pass no question was answered twice, and no answer was dropped, edited or regenerated. Ten of the 500 prompts had each been answered once earlier the same evening in a transport pilot on the same route; the pass answered them again, and the pilot answers are not scored here.

\textbf{Judging.} The saved answers were scored with the released harness and templates. Every call resolved to the snapshot \texttt{gpt-4o-2024-08-06}. Scoring ran in two batches in question-identifier order, 225 and 275 rows, each preceded by its own controls: 17/17 and 20/20 positive, 20/20 and 20/20 negative. The first batch was scored while the pass was still running. A stopping rule had been set before that batch (stop the pass if it trailed the headline pass with exact McNemar $p<0.05$); it did not trigger, and no answer changed. The second batch stopped after 199 rows when the API account's prepaid credit ran out and resumed from the next row; no row was judged twice. The 72 answerable knowledge-update answers were also scored under the exact upstream template (\S5.6), with controls 12/12 and 12/12. As a second judge, all 500 answers and the 500 saved Opus pass-1 answers were scored by Claude Sonnet 5.5 with the same templates as they landed, with no judge error. That judge is not a like-for-like copy of the first: it ran as stateless subscription calls with no tools, a one-line grader system prompt, no temperature control, and medium reasoning effort, a setting chosen after a 99-answer calibration against the GPT-4o verdicts on Opus pass-1 answers. Thirty answers per side were judged twice by it after a watcher restart; the first verdict is the one scored, none of the 30 repeated Sol verdicts differ from the first, and 2 of the 30 repeated Opus verdicts do. Its controls, of the same design as the first judge's, were run after its judgments and passed 19/19 positive and 20/20 negative. It shares a vendor with the headline reader, and no verdict here was checked by a human.

\textbf{Arm A.} The arm-A pass (\S5.8) used the same route, reader line and settings. Its checks differed in two ways: an audit alert could stop the reader, and the token check became the exact count after the first stop described below. Its 494 new calls used 12,839,717 input tokens, none cached, and 46,685 output tokens, of which 18,645 were reasoning tokens (nonzero on 311 calls). The exact token count holds on all 494 calls with the same constant, 688. The pass stopped twice and was resumed: once when the earlier token-count fit raised a false alarm on a token-dense prompt, which is what led to the exact count, and once at a preset account-usage ceiling with three calls left. Two questions were asked twice as a mechanical consequence of those stops. In one the stop arrived after the call had completed and before its record was written; in the other the completed call was flagged only for the usage reading. In both the second answer is the one scored, the first is kept unscored, and the two give the same answer. Neither repeat was a choice made after reading an answer, and no other answer was dropped, edited or regenerated. The GPT-4o judge scored the 494 new answers in one batch after the pass, with controls 19/19 positive and 20/20 negative, and the six reused arm-D answers keep their arm-D verdicts. The 71 new answerable knowledge-update answers score 70 under both the released and the upstream template (controls 12/12 and 12/12). The second judge scored the 494 new answers as they landed, with no judge error. Arm D's calls ran from 03:06 to 04:20 UTC and arm A's from 04:42 to 06:01 UTC, on a model slug that is not pinned.

\section*{References}
\begin{itemize}\setlength{\itemsep}{1pt}\setlength{\parskip}{0pt}\setlength{\parsep}{0pt}
\item Wu, D., et al. \emph{LongMemEval: Benchmarking Chat Assistants on
Long-Term Interactive Memory.} ICLR 2025, arXiv:2410.10813.
\item Sen, S., Lumer, E., Gulati, A., Subbiah, V.~K. \emph{Chronos:
Temporal-Aware Conversational Agents with Structured Event Retrieval for
Long-Term Memory.} arXiv:2603.16862, Mar 2026.
\item Wu, M., Zhu, P. \emph{Agent Zero Memory: Provenance-Aware Long-Term
Memory for LLM Agents.} arXiv:2608.29606, Aug 2026.
\item Barnes, T. \emph{Observational Memory: 95\% on LongMemEval.} Mastra
Research, Feb 2026 (accessed 2026-08-31).
\item Packer, C., et al. \emph{MemGPT: Towards LLMs as Operating Systems.}
arXiv:2310.08560, 2023.
\item Chhikara, P., et al. \emph{Mem0: Building Production-Ready AI Agents with Scalable Long-Term Memory.} arXiv:2504.19413, 2025. \quad Rasmussen, P., et al. \emph{Zep: A Temporal Knowledge Graph Architecture for Agent Memory.} arXiv:2501.13956, 2025.
\item Maharana, A., et al. \emph{Evaluating Very Long-Term Conversational Memory of LLM Agents} (LoCoMo). arXiv:2402.17753, 2024.
\item Nogueira, R., Cho, K. \emph{Passage Re-ranking with BERT.} arXiv:1901.04085, 2019. \quad
Chen, J., et al. \emph{BGE M3-Embedding.} arXiv:2402.03216, 2024. \quad Cormack, G., et al.
\emph{Reciprocal Rank Fusion.} SIGIR 2009.
\item Wang, X., et al. \emph{Self-Consistency Improves Chain of Thought
Reasoning.} ICLR 2023. \quad Zheng, L., et al. \emph{Judging LLM-as-a-Judge.}
NeurIPS 2023. \quad Dodge, J., et al. \emph{Show Your Work.} EMNLP 2019.
\item BAAI. \emph{bge-reranker-v2-m3} (cross-encoder reranker used in the
retrieval chain). \url{https://huggingface.co/BAAI/bge-reranker-v2-m3} (accessed 2026-09-24).
\item OMEGA. \emph{LongMemEval benchmark results.} \url{https://omegamax.co/benchmarks} (accessed 2026-09-24).
\item McCann, J. \emph{agentmemory V4.} GitHub repository, \url{https://github.com/JordanMcCann/agentmemory} (accessed 2026-09-24 and 2026-09-28).
\end{itemize}

\end{document}